%% file: main.tex
\documentclass[letterpaper, 10 pt, conference]{ieeeconf}  

\IEEEoverridecommandlockouts                              

\input{imports}


\input{title}

\begin{document}

\maketitle
\thispagestyle{empty}
\pagestyle{empty}


\input{sections/abstract}


\input{sections/introduction}


\input{sections/related_work}


\input{sections/problem_statement}


\input{sections/technical_approach}


\input{sections/experiments}


\input{sections/results}


\input{sections/conclusion}

\bibliographystyle{IEEEtran}
\bibliography{kelvin_references}

\end{document}

%% file: imports.tex
\usepackage{comment}
\usepackage{amsmath}
\usepackage[dvipsnames]{xcolor}
\usepackage{xspace}
\usepackage{algorithm}
\usepackage{algpseudocode}
\usepackage{graphicx}
\usepackage{svg}
\usepackage{gensymb}
\usepackage{ragged2e}
\usepackage{amssymb}
\usepackage{cite}
\usepackage{soul}
\usepackage[pdfa,colorlinks,bookmarksopen,bookmarksnumbered,allcolors=blue]{hyperref}
\usepackage{subcaption}
\usepackage{multirow}
\usepackage[normalem]{ulem} 

%% file: title.tex
\title{\LARGE \bf
Anticipatory Gait Switching for Quadrupeds via Look-Ahead MDP Planning
}

\author{
Kelvin Aladum\textsuperscript{1}, Justin Peel\textsuperscript{2}, Nisar Ahmed\textsuperscript{1}
}

%% file: sections/abstract.tex
\begin{abstract}
Quadruped robots often switch gaits to traverse mixed terrain safely. Many practical supervisors enforce hardware constraints (e.g., dwell or guard intervals) using a transition handler. However, the switch decision itself is often driven by near-field thresholds, which can delay commitment when rough terrain is only detected shortly before contact in the local height map.
We propose a lightweight height-map-only gait supervisor that unifies look-ahead and locomotion constraints in a single decision policy. A robot-centered height map is partitioned into near-, mid-, and far-field bands, each discretized into a terrain class, and band evolution is modeled as a Markov process. Gait selection is cast as a look-ahead Markov Decision Process whose transition model and cost explicitly encode platform-specific switching constraints.
Indoor and outdoor experiments on a Unitree Go2 platform show reduced late switching, lower time spent using ineffective gaits on rough terrain, and improved roll stability compared to tuned reactive baselines. Overall, the MDP formulation yields anticipative constraint-aware gait switching that improves robustness without course-specific retuning.
\end{abstract}

%% file: sections/introduction.tex
\section{Introduction} \label{section_introduction}
Legged robots often operate over terrain that changes within a few steps, from flat ground to curbs, stairs, rubble, or uneven outdoor surfaces. Quadrupeds handle this variability by switching among locomotion gaits with different trade-offs in speed, stability, and clearance. Trotting is efficient on smooth ground but can become risky on rough terrain due to foot scuffing and body impacts. Climbing or stair gaits improve clearance and stability but are slower and inefficient when used unnecessarily.

A central challenge is not only how robots should walk, but when their manner of walking should change. Existing systems either rely on rich perception and learned policies for terrain-aware locomotion \cite{tsounis2020deepgait}, or use perception-driven heuristics that trigger gait switching from local terrain cues \cite{havoutis2013onboard}. Here, reactive switching refers to gait switching decisions driven by instantaneous local cues rather than look-ahead, cost-aware planning. On physical platforms, reactive switching can be unreliable when gait transitions are disruptive or delayed by onboard controller logic and safety checks \cite{zimmermann2021go}.

This issue is pronounced on the Unitree Go2 studied in this work, where gait switching is asymmetric. Switching from climb to trot is essentially seamless. In contrast, switching from trot to climb requires the robot to stop and wait for the gait to engage. If it does not, the transition can be poorly handled and lead to stumbling. In some cases, the command is ignored and the robot continues trotting.
Simple reactive threshold rules can enforce conservative stop-and-wait behavior, but they do not decide when it is worth paying the transition cost based on what lies ahead. Height maps from low-cost LiDAR can also be noisy and can overestimate obstacles in vegetation such as tall grass, which can trigger unnecessary switching to climb gait.

To address this, we propose a lightweight gait switching supervisor for LiDAR-enabled quadrupeds that uses an instantaneous robot-centered height map and makes gait switching decisions using a Markov Decision Process (MDP). Fig.~\ref{fig:overview} provides an overview of the proposed pipeline. The forward height map is partitioned into distance bands that provide a compact look-ahead representation. Each band is summarized into a discrete terrain class using robust elevation statistics, and uncertainty in the band evolution is captured in the transition model. The resulting MDP explicitly accounts for asymmetric switching latency and produces an offline-computed policy for online gait switching.
\input{Figures/front_page}

The main contributions of this work are:
\begin{itemize}
    \item A height map only, anticipative gait switching supervisor that requires minimal sensing and is compatible with existing gait controllers;
    \item An MDP formulation that models asymmetric gait transition latency, including a dwell state for trot-to-climb switching, and yields a compact optimal switching policy solved offline;
    \item An empirical indoor and outdoor evaluation on the Unitree Go2 comparing the proposed look-ahead MDP policy against a tuned hysteresis baseline, demonstrating improved switching behavior under real transition constraints.
\end{itemize}
We describe related work and our problem formulation in Sec.~\ref{section_related_work_probst}. Sec.~\ref{subsec:terrain} describes the height map processing and band abstraction technique and Sec.~\ref{subsec:mdp} presents the MDP model and offline optimal solution procedure. Secs.~\ref{section_experiments} and \ref{section_results} evaluate the approach in representative mixed-terrain experiments.

%% file: Figures/front_page.tex
\begin{figure}[t!]
    \centering
    \includegraphics[width=\columnwidth]{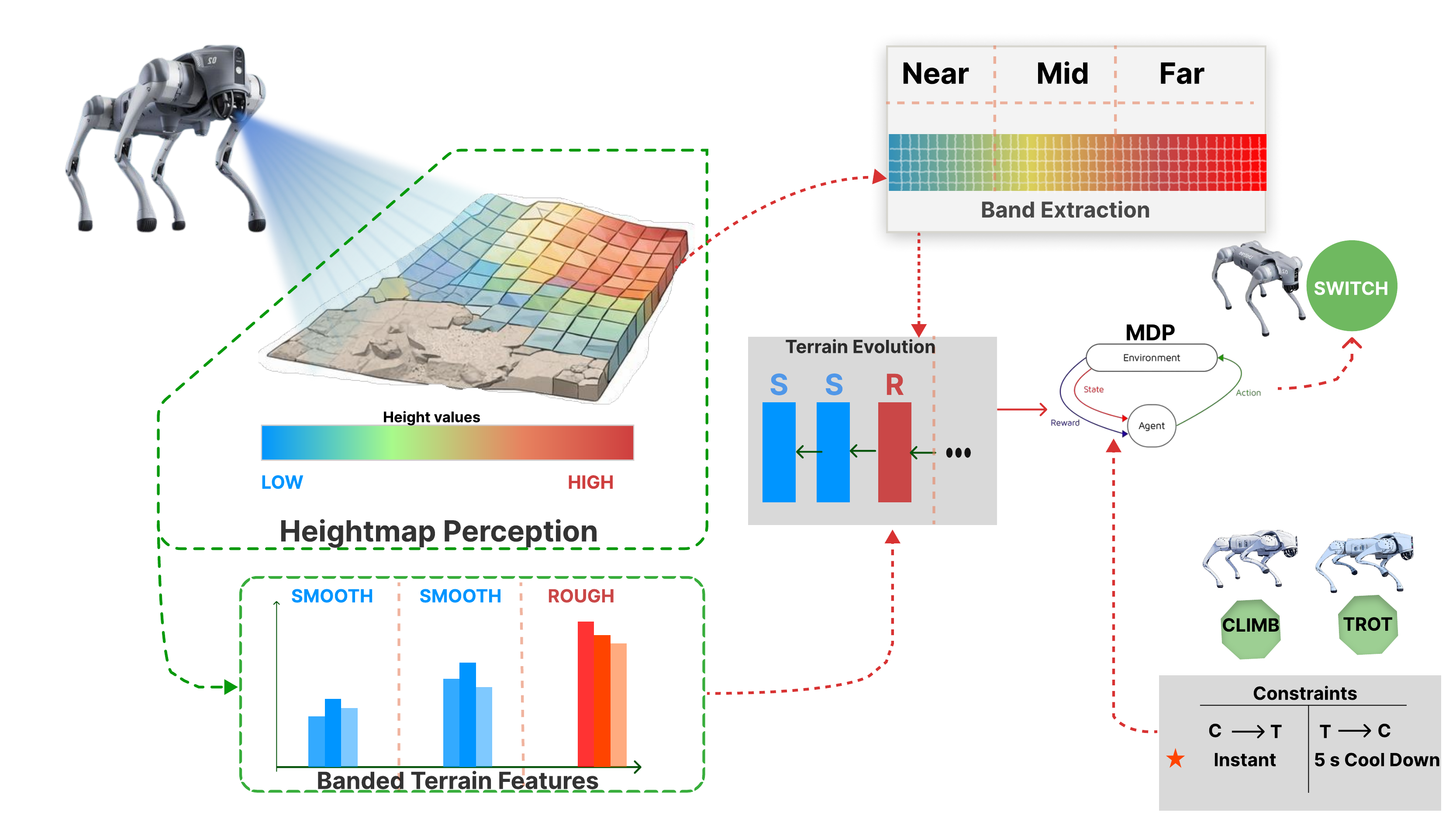}
    \label{fig_lsp_planning}
    \caption{Banded look-ahead from a local height map enables anticipative trot–climb switching.}
    \label{fig:overview}
\end{figure}

%% file: sections/related_work.tex
\section{Related Work and Problem Statement} \label{section_related_work_probst}
Gait selection for quadruped robots has been approached through reactive threshold methods, optimization-based frameworks, and learning-based policies. Early work relied on manually tuned finite state machines that switch gaits based on velocity or terrain features extracted from elevation maps \cite{koo2015biologically, fankhauser2018robust}. While computationally efficient, these reactive approaches do not explicitly reason about upcoming terrain and often require substantial retuning across platforms and controllers. Central Pattern Generator (CPG) based methods \cite{shang2019smooth,li2016gait} achieve smooth gait generation and transitions through coupled oscillators, but the switching logic is typically heuristic, for example triggered by stability metrics such as the zero moment point, rather than derived from an explicit decision objective.

Optimization-based approaches formulate gait switching as an offline or online planning problem. Wang et al. \cite{wang2020automatic} pioneered automatic gait pattern selection by using MINLP to solve trajectory optimization across different gaits and task conditions, then compressing the solution space via supervised learning. While this reduces manual tuning, their formulation optimizes full-body trajectories and contact sequences rather than the discrete switching decision of when to change gaits, and it does not explicitly model direction-dependent transition latency or terrain look-ahead. Saraf et al. \cite{saraf2021terrain} developed a Model Predictive Control (MPC) approach that explicitly plans transitions between bounding and trotting to enhance stability, but the online optimization budget limits the effective prediction horizon to a few seconds and makes gait switching tightly coupled to low-level control. Recent work by Kotecha et al. \cite{kotecha2025real} combines a model predictive path integral (MPPI) controller with learned abstractions to jointly optimize actions and continuous gait parameters, improving energy efficiency compared to baselines, but requires computationally expensive online optimization at each control step.

Deep reinforcement learning has enabled end-to-end learning of gait switching policies. Shafiee et al. \cite{shafiee2024viability} used viability (i.e., avoiding falls) as the primary criterion for emergent gait transitions in their bio-inspired Deep RL framework, with energy efficiency as a secondary objective. Bellegarda and Humphreys \cite{bellegarda2025allgaits,humphreys2025learning} trained a dedicated switching policy that minimizes bio-inspired transition metrics, including Froude number and duty factor, achieving robust real-world performance on terrains similar to those seen during training. Yang et al. \cite{yang2022fast} used evolutionary strategies to learn hierarchical gait policies that automatically transition between walking, trotting, and fly-trotting based on speed. While these learning-based methods can achieve strong performance within their training distribution, they often provide limited interpretability, require extensive simulation training, and can require retraining to adapt to new scenarios, hardware constraints, or gait types not represented during training. Recent multi-gait frameworks \cite{defazio2024learning} learn diverse gait modules through CPG integration or latent skill representations but commonly assume a high-level gait command, rather than autonomous terrain-triggered switching under controller constraints.

Energy and stability criteria have been incorporated into gait mapping approaches. Liu et al. \cite{zhang2024novel} collected hundreds of locomotion trials measuring cost of transport and stability margins across speeds and terrains, then constructed offline lookup tables for gait choice with FSM-managed transitions. Their Unitree Go2 experiments demonstrated terrain-dependent preferences across gaits, but the approach does not model transition costs, does not represent look-ahead terrain evolution, and leaves the switching timing to a hand-designed transition manager.

Our work differs from these approaches in three key aspects. First, we formulate gait switching as an MDP with a compact, banded representation of the height map that encodes look-ahead terrain evolution, enabling anticipatory switching decisions. Second, we incorporate non-ideal transition behavior typical of real legged hardware, including latency, stabilization, and direction-dependent switching overhead, directly into the decision model via explicit transition states. Third, unlike online optimization methods (MPC, MPPI) or learned policies (Deep RL), our offline value iteration solution deploys as a small lookup table with minimal runtime overhead, while maintaining interpretability and adaptability via MDP re-solving rather than retraining. In contrast to general anticipatory online planning methods such as \cite{burns2012anticipatory}, which perform online look-ahead search over abstract action models, our method uses a stochastic model of terrain class evolution derived from the robot's height map abstraction and explicitly captures gait-transition constraints imposed by the onboard controller.

%% file: sections/problem_statement.tex
\subsection{Problem Statement} \label{subsection_problem_statement}
We study high-level gait switching for a quadruped using a robot-centered, short-horizon 2.5D LiDAR height map observed online in previously unseen environments. The robot may support multiple locomotion modes with direction-dependent transition requirements; in this work, we focus on switching between a fast trot ($T$) and a cautious climb ($C$) to isolate the decision problem under real transition constraints. At discrete decision times $t=0,1,2,\dots$ with period $\Delta t$, the supervisor must choose whether to remain in the current gait or initiate a switch.

On real hardware, gait transitions can be asymmetric and delayed by controller logic and safety checks. In particular, the $T\rightarrow C$ transition may require stopping and dwelling before the new gait reliably engages. The objective is to compute a switching strategy that anticipates rough terrain early enough to complete this transition cost, while avoiding unnecessary switching on smooth ground despite noisy height maps and short look-ahead.
\input{Figures/system_architect}

%% file: Figures/system_architect.tex
\begin{figure}[t]
  \centering
\includegraphics[width=\linewidth]{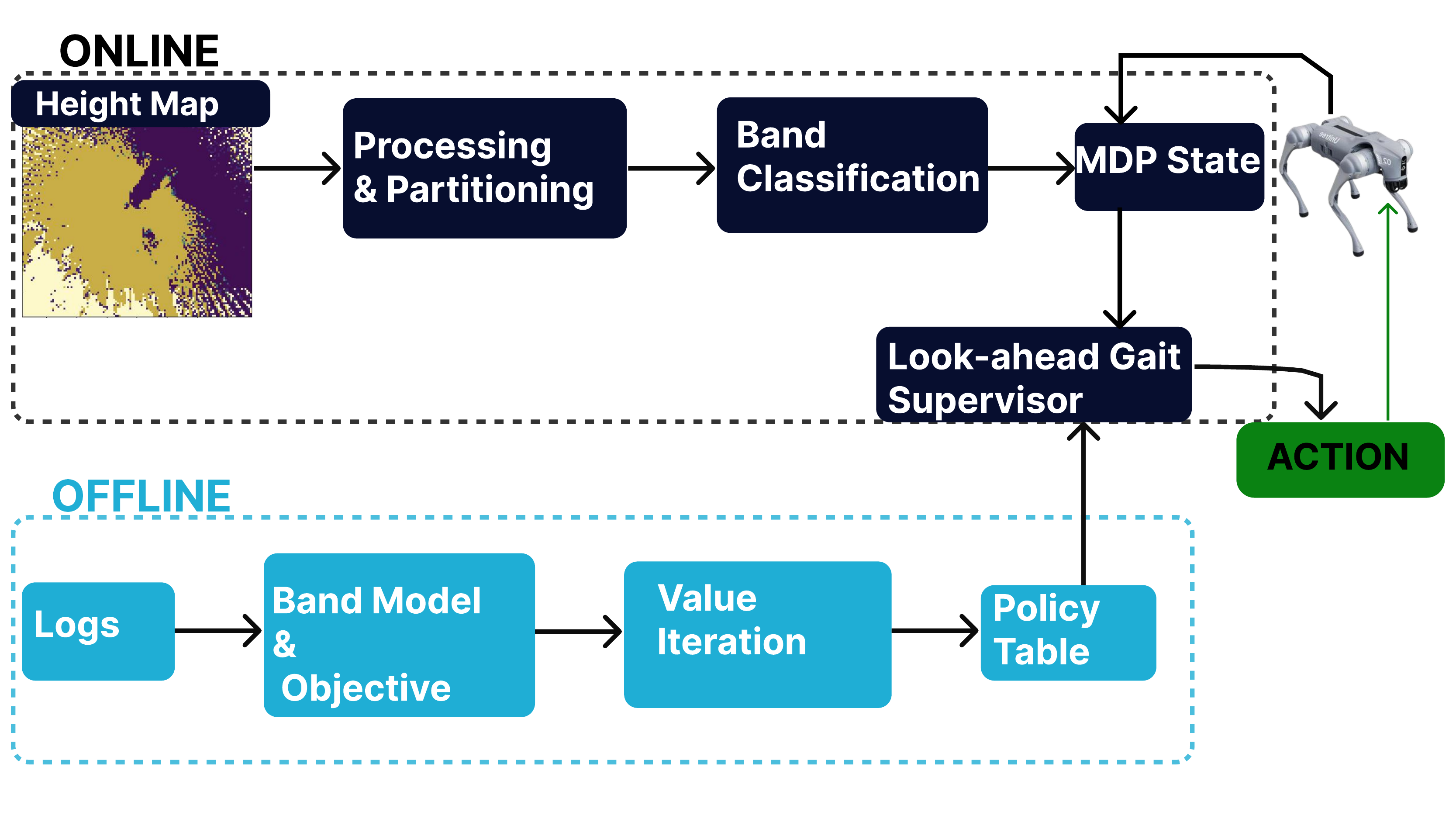}
  \caption{System architecture for height-map-only gait switching.}
  \label{fig:system_architecture}
\end{figure}

%% file: sections/technical_approach.tex
\section{Methodology} \label{section_methodology}

\subsection{System Overview} \label{subsec:system_overview}

Fig.~\ref{fig:system_architecture} summarizes the proposed height map-only gait supervisor. At each decision step, the incoming height map is filtered for invalid cells and cropped to a forward region aligned with the robot heading. The forward region is partitioned into range-ordered distance bands, and each band is summarized with robust elevation statistics that are discretized into terrain classes. These classes, together with the current gait and any active transition progress, form a compact MDP state.

The MDP is solved offline, and the resulting stationary policy is exported as a small lookup table. Online, the supervisor evaluates this table at a fixed period of $\Delta$t and outputs either \textsc{Stay} or \textsc{Switch}. For example, if the near band is smooth but the mid and far bands are classified as rough while the robot is in $T$, the policy can return \textsc{Switch} to start the $T\rightarrow C$ procedure early enough for the dwell to complete. Online execution applies the MDP-defined mode update and issues the corresponding gait API commands. In particular, when the policy selects \textsc{Switch} in $T$, the supervisor transitions into the dwell mode $S$ and enforces the required stop-and-dwell before entering $C$.

\subsection{Terrain Representation from Heightmaps} 
\label{subsec:terrain}

\subsubsection{Height Map Pre-processing and Valid Cells}
Let $Z \in \mathbb{R}^{H \times W}$ denote the incoming robot-centered 2.5D height map, where
$H$ and $W$ are the grid height and width, and each cell stores an elevation $Z_{ij}$ in meters at resolution $r$ (m/cell). Invalid cells are identified by a sentinel value $Z_{\mathrm{inv}}$. We define a binary validity mask for indices $i \in \{1,\dots,H\}$ and $j \in \{1,\dots,W\}$:
\begin{equation}
\mathcal{V}(i,j) =
\mathbb{I}\!\left[\mathrm{isfinite}(Z_{ij}) \ \wedge\  Z_{ij} \neq Z_{\mathrm{inv}}\right],
\end{equation}
where $\mathrm{isfinite}(\cdot)$ returns true for finite real values, and $\mathbb{I}[\cdot]$ is the indicator function.

If the number of valid cells inside the forward region of interest is below a minimum count $N_{\min}$, the supervisor abstains from switching and returns \textsc{Stay}. Here $N_{\min}$ is a fixed robustness parameter chosen to prevent decisions from being made on sparse or unreliable height maps; its value is set once and reported with other implementation parameters.

\subsubsection{Forward Wedge and Local Ground Reference}

Each cell $(i,j)$ is mapped to a robot-centric planar coordinate $(x_{ij},y_{ij})$ with the grid origin at the map center. Let $\psi$ denote the robot yaw. We compute the forward-relative bearing and range
$
\theta_{ij}=\mathrm{wrap}\!\left(\mathrm{atan2}(y_{ij},x_{ij})-\psi\right), 
\qquad
R_{ij}=\sqrt{x_{ij}^2+y_{ij}^2},$
where $\mathrm{wrap}(\cdot)$ maps angles to $(-\pi,\pi]$. The forward wedge region of interest is
\begin{equation}
\mathcal{W}=\left\{(i,j):\ \mathcal{V}(i,j)=1,\ |\theta_{ij}|\le\theta_{\max},\ R_{ij}\le R_{\max}\right\}.
\label{eq:wedge}
\end{equation}

Outdoor runs revealed that ground estimates can be biased when the wedge contains negative obstacles, i.e., local terrain that drops below the current support surface (such as curbs and ditches). To reduce this sensitivity, we estimate a local ground reference from a short-range front patch aligned with the robot heading:
\begin{equation}
\mathcal{G}_0=\left\{(i,j):\ \mathcal{V}(i,j)=1,\ |\theta_{ij}|\le\theta_g,\ R_{ij}\in [R_g^{\min},R_g^{\max}]\right\},
\end{equation}
with $\theta_g < \theta_{\max}$. If $|\mathcal{G}_0|<N_{\min}$, we expand to wider fallback patches $\mathcal{G}_1,\mathcal{G}_2$ (defined by increasing $\theta_g$ and range) until enough valid cells are available. Let $\mathcal{G}$ denote the first patch in this sequence with at least $N_{\min}$ valid cells.

Ground height is estimated as a low percentile
\begin{equation}
z_g = \mathrm{Percentile}\!\left(\{Z_{ij}:(i,j)\in\mathcal{G}\},\,p_g\right),
\label{eq:ground}
\end{equation}
and relative heights are defined as $Z^{\mathrm{rel}}_{ij}=Z_{ij}-z_g$ for subsequent feature computation.

\subsubsection{Look-Ahead Banding and Feature Extraction}

To provide look-ahead information, we partition the forward wedge $\mathcal{W}$ into $K$ radial bands indexed by $k\in\{0,\dots,K-1\}$ using predefined range limits $\{[R_k^{\min},R_k^{\max})\}$:
\begin{equation}
\mathcal{B}_k=\left\{(i,j)\in\mathcal{W}:\ R_{ij}\in [R_k^{\min},R_k^{\max})\right\}.
\end{equation}

For each band $\mathcal{B}_k$, we compute robust statistics from the relative elevations $\{Z^{\mathrm{rel}}_{ij}:(i,j)\in\mathcal{B}_k\}$:
(i) a high-percentile height $z_{k}^{p_h}$, (ii)an interquartile range $\mathrm{IQR}_{k}$, and
(iii) a high-percentile slope magnitude $s_{k}^{p_s}$. The slope magnitude is computed from finite differences on the relative-height grid,
$\|\nabla Z^{\mathrm{rel}}_{ij}\|$, using resolution $r$ for scaling.
These features summarize obstacle presence (high percentiles), roughness (IQR), and local steepness (slope).

Outdoor tests also showed that low-elevation hazards can be underrepresented by high-percentile statistics. We therefore add a per-band \emph{drop} signal based on a low percentile of relative height:
\begin{equation}
\begin{aligned}
p_{k}^{p_d}&=\mathrm{Percentile}\!\left(\{Z^{\mathrm{rel}}_{ij}:(i,j)\in\mathcal{B}_k\},\,p_d\right), \\
d_{k}^{\mathrm{drop}}&=\max(0,\,-p_{k}^{p_d}),
\end{aligned}
\end{equation}
where $p_d$ is a fixed low percentile. Note that $Z^{\mathrm{rel}}_{ij}$ can be negative when a region lies below the local ground estimate $z_g$, so $p_{k}^{05}$ can be negative and $d_{k}^{\mathrm{drop}}$ need not be zero.

Finally, we compute a wedge-level grade signal to capture sustained uphill or downhill structure. Let
\begin{align*}
\tilde{v}_{k}=\mathrm{Median}\!\left(\{Z^{\mathrm{rel}}_{ij}:(i,j)\in\mathcal{B}_k\}\right), \\
\bar{R}_k=\tfrac{1}{2}(R_k^{\min}+R_k^{\max}),
\end{align*}
and define
\begin{equation}
g=\frac{\tilde{v}_{k_f}-\tilde{v}_{k_n}}{\bar{R}_{k_f}-\bar{R}_{k_n}},
\end{equation}
where $k_n$ and $k_f$ denote the chosen near and far bands. We use $|g|$ as the grade magnitude for discretization.

\subsubsection{Band Discretization into Terrain Classes} 
To keep the decision model compact, each band is mapped to a discrete terrain class
$c_k \in \{0,1\}$ denoting \textsc{Smooth} ($0$) or \textsc{Rough} ($1$). We normalize each scalar feature $x$ to $[0,1]$ using a clamped linear map
\begin{equation}
\mathrm{norm}(x)=\mathrm{clip}\!\left(\frac{x-x_T}{x_C-x_T},\,0,\,1\right),
\label{eq:norm}
\end{equation}
where $\mathrm{clip}(\cdot,0,1)$ clamps to $[0,1]$, and $x_T$ and $x_C$ are fixed normalization endpoints chosen offline from calibration logs and then held constant for all runs. For a given feature $x$, let $\mathcal{D}_T(x)$ denote the set of recorded values of $x$ on segments labeled \textsc{Smooth} during steady trotting, and let $\mathcal{D}_C(x)$ denote the set of recorded values of $x$ on segments labeled \textsc{Rough} during steady climbing. We set
\begin{equation}
\begin{aligned}
x_T=\mathrm{Percentile}\!\big(\mathcal{D}_T(x),\,p_T\big),\\
x_C=\mathrm{Percentile}\!\big(\mathcal{D}_C(x),\,p_C\big),
\end{aligned}
\end{equation}
where $\mathrm{Percentile}(\cdot,p)$ returns the $p$-th percentile of its input set and $p_T,p_C\in(0,100)$ are fixed choices. 

We compute three difficulty scores: (i) a three-feature roughness score,
\begin{equation}
d^{3}_k = \tfrac{1}{3}\Big(\mathrm{norm}(z_{k}^{p_h}) + \mathrm{norm}(\mathrm{IQR}_{k}) + \mathrm{norm}(s_{k}^{p_s})\Big),
\end{equation}
(ii) a drop score, $d^{\mathrm{drop}}_k=\mathrm{norm}(d_{k}^{\mathrm{drop}})$, and
(iii) a grade score $d^{\mathrm{grade}}=\mathrm{norm}(|g|)$ (shared across bands). The final band difficulty is taken as a conservative fusion:
\begin{equation}
d_k = \max\!\Big(d^{3}_k,\ d^{\mathrm{drop}}_k,\ d^{\mathrm{grade}}\Big),
\qquad
c_k = \mathbb{I}[d_k \ge \tau],
\end{equation}
where $\tau$ is a fixed threshold. The observation at time $t$ is the band-class tuple $\mathbf{c}_t= (c_{0,t},\dots,c_{K-1,t})$ for $K$ bands.

\begin{algorithm}[t]
\caption{Online Height-map-only MDP Gait Supervisor}
\label{alg:online_supervisor}
\begin{algorithmic}[1]
\Require Height map $Z$, yaw $\psi$, current mode $m_t\in\{T,C,S\}$, dwell counter $\delta_t$, policy table $\pi(\cdot)$.
\Require Parameters: $\theta_{\max}$, $R_{\max}$, band ranges $\{[R_k^{\min},R_k^{\max})\}$, $N_{\min}$, dwell ticks $D$.
\State Build valid mask $\mathcal{V}$; construct forward wedge $\mathcal{W}$ using $(\psi,\theta_{\max},R_{\max})$
\If{$|\mathcal{W}| < N_{\min}$} \State \Return \textsc{Stay} \EndIf
\State Estimate local ground $z_g$ from $\mathcal{G}$ and compute relative heights $Z^{\mathrm{rel}}_{ij}=Z_{ij}-z_g$
\State Compute grade magnitude $|g|$ from chosen near/far band medians
\For{$k=0$ to $K-1$} 
  \State Extract $\mathcal{B}_k$ and compute $(z^{(k)}_{95}, \mathrm{IQR}^{(k)}, s^{(k)}_{90}, d^{(k)}_{\mathrm{drop}})$
  \State Discretize to class $c_{k,t}\in\{0,1\}$ using $d_k=\max(\cdot)$ and threshold $\tau$
\EndFor
\State Form state $s_t=(m_t,\delta_t,\mathbf{c}_t)$ and query action $a_t \leftarrow \pi(s_t)$
\If{$a_t=\textsc{Stay}$}
  \State \Return \textsc{Stay}
\EndIf
\Comment{\textsc{Switch} is interpreted by a transition handler based on the current mode}
\If{$m_t=T$} \Comment{$T\rightarrow C$ requires stop-and-dwell}
  \State Command zero velocity
  \State Set $m_t \leftarrow S$, $\delta_t \leftarrow D$
\ElsIf{$m_t=C$} \Comment{$C\rightarrow T$ is immediate}
  \State Set $m_t \leftarrow T$, $\delta_t \leftarrow 0$
\Else \Comment{$m_t=S$: dwell in progress}
  \State $\delta_t \leftarrow \delta_t - 1$
  \If{$\delta_t = 0$}
    \State Set $m_t \leftarrow C$
  \EndIf
\EndIf
\State \Return \textsc{Switch}
\end{algorithmic}
\end{algorithm}

\subsection{MDP Formulation for Transition-Aware Gait Switching}
\label{subsec:mdp}
We model gait switching as a \emph{finite-state, infinite-horizon} Markov Decision Process (MDP) defined by the tuple $(\mathcal{S},\mathcal{A},P,\mathcal{R},\gamma)$, where $\mathcal{S}$ is a finite set of states, $\mathcal{A}$ is a finite set of actions, $P(s'|s,a)$ is the transition model, $\mathcal{R}(s,a)$ is the reward, and $\gamma\in(0,1)$ is the discount factor.

\subsubsection{State and Actions}
The finite state space is
\begin{equation}
\mathcal{S}=\mathcal{M}\times\{0,1,\dots,D\}\times\mathcal{C}^K,
\end{equation}
where $\mathcal{M}=\{T,C,S\}$ denotes the locomotion mode (trot, climb, or $T\!\rightarrow\!C$ dwell), $\mathcal{C}=\{0,1\}$ denotes the per-band terrain class set (\textsc{Smooth}/\textsc{Rough}).

At decision step $t=0,1,2,\dots$, the MDP state variable $s_t\in\mathcal{S}$ takes the value
\begin{equation}
s_t \triangleq (m_t,\delta_t,\mathbf{c}_t),
\end{equation}
where $m_t\in\mathcal{M}$, $\delta_t\in\{0,1,\dots,D\}$ is the dwell counter (nonzero only when $m_t=S$), $D$ is the number of dwell ticks, and $\mathbf{c}_t\in\mathcal{C}^K$ is the band-class tuple.

The action set is $\mathcal{A}=\{\textsc{Stay},\textsc{Switch}\}$, where \textsc{Switch} is interpreted as a mode-dependent toggle.

\subsubsection{Mode Transition Dynamics (Asymmetric Switching)}
Let $T_{\mathrm{dwell}}$ denote the required stop-and-dwell duration for reliable $T\!\rightarrow\!C$ engagement, and let $\Delta t$ be the decision period. We define the number of dwell ticks as
\begin{equation}
D=\left\lceil \frac{T_{\mathrm{dwell}}}{\Delta t}\right\rceil
\end{equation}
The deterministic mode update is:
\begin{itemize}
  \item If $m_t=T$ and $a_t=\textsc{Switch}$: set $m_{t+1}=S$ and $\delta_{t+1}=D$.
  \item If $m_t=C$ and $a_t=\textsc{Switch}$: set $m_{t+1}=T$ and $\delta_{t+1}=0$.
  \item If $m_t=S$: action is ignored; if $\delta_t>1$, set $m_{t+1}=S$ and $\delta_{t+1}=\delta_t-1$; otherwise set $m_{t+1}=C$ and $\delta_{t+1}=0$.
  \item Otherwise (\textsc{Stay} in $T$ or $C$): set $m_{t+1}=m_t$ and $\delta_{t+1}=0$.
\end{itemize}

\subsubsection{Terrain Band Dynamics (Look-Ahead Model)}
When the robot is moving ($m_t\in\{T,C\}$), we approximate look-ahead progression by shifting the band classes:
\begin{equation}
c_{0,t+1}=c_{1,t}, \qquad c_{1,t+1}=c_{2,t},
\end{equation}
and evolving the far band via a two-state Markov model
\begin{equation}
\Pr(c_{2,t+1}=j \mid c_{2,t}=i)=\mathbf{P}_{\mathrm{far}}[i,j], \qquad i,j\in\{0,1\},
\end{equation}
where $\mathbf{P}_{\mathrm{far}}$ is estimated from logged far-band class sequences using smoothed frequency counts. When stopped in dwell ($m_t=S$), we approximate $\mathbf{c}_{t+1}=\mathbf{c}_t$.

The full transition model $P(s'|s,a)$ is constructed by combining (i) the deterministic mode update and (ii) the terrain evolution model for $\mathbf{c}_t$ conditioned on whether the mode is moving or dwelling.

\subsubsection{Reward Function}
We define the reward as $\mathcal{R}(s_t,a_t)$ to trade off traversal progress, terrain safety, and switching overhead:
\begin{equation}
\begin{aligned}
\mathcal{R}(s_t,a_t)=
\mathcal{R}_{\text{mode}}(m_t)
- C_{\text{sw}}\,\mathbb{I}[a_t=\textsc{Switch}] \\
+ \mathcal{R}_{\text{terr}}(m_t,c_{0,t}),
\end{aligned}
\end{equation}
where $C_{\text{sw}}>0$ discourages frequent switching.. We use
\begin{equation}
\mathcal{R}_{\text{mode}}(m_t)=
\begin{cases}
\rho_{\text{move}}, & m_t \in \{T,C\},\\
-\,\rho_{\text{dwell}}, & m_t = S,
\end{cases}
\end{equation}
with $\rho_{\text{move}}>0$ rewarding time spent moving and $\rho_{\text{dwell}}>0$ penalizing time spent dwelling. Terrain mismatch is
\begin{equation}
\mathcal{R}_{\text{terr}}(m_t,c_{0,t})=
\begin{cases}
-\,C_{\text{TR}}, & m_t=T \ \land\ c_{0,t}=1,\\
-\,C_{\text{CS}}, & m_t=C \ \land\ c_{0,t}=0,\\
0, & \text{otherwise},
\end{cases}
\end{equation}
where $C_{\text{TR}}\gg C_{\text{CS}}>0$ penalizes trotting on rough near-band terrain more strongly than climbing on smooth terrain.

\subsubsection{Offline Policy Computation and Export}
Because $\mathcal{S}$ and $\mathcal{A}$ are finite and the task is continuing ($t=0,1,2,\dots$), we compute an optimal stationary policy $\pi^*:\mathcal{S}\rightarrow\mathcal{A}$ via value iteration:
\begin{equation}
V^*(s)=\max_{a\in\mathcal{A}}\Big(\mathcal{R}(s,a)+\gamma\sum_{s'\in\mathcal{S}}P(s'|s,a)\,V^*(s')\Big).
\end{equation}
The deployed policy $\pi^*(s)=\arg\max_{a\in\mathcal{A}}(\cdot)$ is exported as a compact lookup table for online evaluation.

\subsection{Online Execution}
\label{subsec:online}

Algorithm~\ref{alg:online_supervisor} summarizes the online supervisor executed at period $\Delta t$. At each step it maps the incoming height map to the band-class tuple $\mathbf{c}_t$, queries the precomputed policy, and applies the MDP mode dynamics.

If the policy selects \textsc{Stay}, the current mode is maintained. If it selects \textsc{Switch}, the supervisor performs the mode update defined in Sec.~\ref{subsec:mdp}: from $C$ it transitions immediately to $T$, and from $T$ it enters the dwell mode $S$ to enforce the required stop-and-dwell before transitioning to $C$. Executing a mode change consists of issuing the corresponding gait command to the robot controller and, when in 
$S$, commanding zero velocity for the dwell duration.

%% file: sections/experiments.tex
\section{Experiments} \label{section_experiments}
\subsection{Platform and Execution Setup}
Experiments were conducted on a Unitree Go2 using the proposed height-map-only switching supervisor. We evaluate switching between trot ($T$) and the platform's climb/stair mode ($C$) \cite{unitree_go2_manual,unitree_ai_motion_service}. The $T\!\rightarrow\!C$ transition on this platform requires a stop-and-dwell; all methods were evaluated under the same measured dwell timing and decision period (Table~\ref{tab:impl_params}) to ensure differences reflect switching decisions rather than transition execution.

To also isolate switching decisions from path following, each run used the same start-to-goal traverse. The robot was commanded to move forward at a fixed speed in each gait (trot and climb), while an operator provided small yaw corrections to maintain an approximately straight heading. The operator did not command gaits; all gait changes were triggered solely by the evaluated switching policies.

\subsubsection{Implementation Parameters}
All fixed parameters used by the perception pipeline and decision model are reported in Table~\ref{tab:impl_params}, including the height map size and resolution, invalid-cell sentinel, wedge geometry, band ranges, ground-estimation parameters, difficulty thresholds, decision period $\Delta t$, and the measured dwell time $T_{\mathrm{dwell}}$ (and corresponding dwell ticks $D$).

\input{Tables/implementation}

\subsection{Test Courses}\label{subsec:course}
We evaluate on three courses shown in Fig.~\ref{fig:courses} (two outdoor and one indoor, with start/goal markers). The courses are chosen to include (i) smooth segments where unnecessary climbing reduces efficiency, and (ii) abrupt terrain changes where late $T\!\rightarrow\!C$ decisions increase risk.

\subsubsection{Course 1: Indoor Lab Obstacles.}
A controlled indoor course assembled from modular obstacles (foam sheets, a low platform/ramp, and a wooden block), producing repeatable height discontinuities and short rough patches.

\subsubsection{Course 2: Outdoor Paver Crossing.}
A traverse beginning on low grass, transitioning onto a short paved section, and terminating in taller grass. This course induces compliance and traction changes and mild grade variations.

\subsubsection{Course 3: Outdoor Sidewalk Crossing.}
A start-to-goal traverse beginning in a vegetation bed, crossing onto a concrete sidewalk, and reaching the goal. This course includes a pronounced surface transition curb and uneven ground that stresses step-up/step-down perception.
\input{Figures/all_courses}
\subsection{Feature Characterization and Look-Ahead}
To motivate the height map statistics and the role of look-ahead banding, we characterize feature behavior on Course~1 snapshots. For each obstacle type (big foam, small foam, flat wood), the robot was held stationary at a fixed stand-off distance and height maps were logged for $\approx 30$\,s. We compute the same banded quantities used online, including the band difficulty $d_k$ (Sec.~\ref{subsec:terrain}).

Fig.~\ref{fig_feat_dist}a shows the empirical CDF of $d_k$ for the near/mid/far bands; the dashed line indicates the discretization threshold $\tau$ (Table~\ref{tab:impl_params}). Because the robot is placed at a stand-off distance, obstacle signatures appear first in the far band, while near and mid bands remain largely below threshold. Fig.~\ref{fig_feat_dist}b and Fig.~\ref{fig_feat_dist}c show that the far-band statistics (e.g., $z^{95}_{k}$, $\mathrm{IQR}_{k}$, and $s^{90}_{k}$) separate obstacle types with consistent margins. Drop and grade terms remain small in these indoor snapshots, consistent with the absence of step-down and sustained slope structure. Overall, the distributions illustrate how banding preserves range ordering, enabling earlier switching decisions when far bands become rough before the near band.

\input{Figures/feature_char}

\subsection{Compared Supervisors}\label{subsec:supervisors}
We compare three switching supervisors. All methods share the same sensing and preprocessing pipeline (forward-wedge extraction, band features, and terrain discretization) and the same execution of the platform transition constraints. They differ only in how they map extracted signals to \textsc{Stay}/\textsc{Switch} decisions:

\subsubsection{RFT (Reactive Feature Thresholding).}
A hand-tuned baseline that triggers switching using near-band feature thresholds with different enter/exit values (hysteresis). This represents common finite-state/threshold switching practice in legged systems \cite{havoutis2013onboard}. 

\subsubsection{DCH (Difficulty-Class Hysteresis).}
A reactive baseline that collapses near-band terrain into a scalar difficulty $d_0$ and applies hysteresis on $d_0$ to classify the near band as \textsc{Smooth} or \textsc{Rough}; the gait is set to trot on \textsc{Smooth} and climb on \textsc{Rough}. Discretizing perceived terrain into a small number of classes to support gait selection is common in legged locomotion \cite{zenker2013visual}.

\subsubsection{LA-MDP (Look-Ahead MDP Policy).}
Our proposed anticipatory policy that uses the full band tuple (near/mid/far), the current mode, and the transition progress, and queries an offline-computed MDP policy (table lookup) to select \textsc{Stay} or \textsc{Switch}.

\subsection{Protocol}
For each course and each method, we report metrics over 10 completed start-to-goal traversals. For safety, if a trial became non-recoverable (e.g., the robot tripped or became stuck), the operator terminated the attempt and restarted until a completed traversal was obtained. These terminated attempts were not recorded as explicit failure labels in the logs; therefore, the quantitative results in Sec.~\ref{section_results} summarize behavior conditional on completion.
To still capture safety-relevant behavior, we additionally report risk and stability proxies derived from onboard logs (trot-on-rough exposure, dwell overhead, and roll). All methods used the same decision period $\Delta t$ and the same measured $T\!\rightarrow\!C$ dwell timing (Table~\ref{tab:impl_params}).

\subsection{Metrics}
From per-tick logs (mode $m_t$, decisions $a_t$, band classes $\mathbf{c}_t$, dwell time in $S$, commanded motion, and body orientation), we compute:

\subsubsection{Switch count}
Total number of realized gait changes per run, computed from discrete changes in the logged mode.

\subsubsection{Late $T \rightarrow C$ switch rate}
Fraction of $T\!\rightarrow\!C$ switches initiated after the near band is already \textsc{Rough} (i.e., after $c_{0,t}$ first becomes $1$). Lower is better.

\subsubsection{Switch anticipation distance}
Signed distance between the $T\!\rightarrow\!C$ initiation point and the near-band rough-entry point. Negative values indicate switching before near-band rough entry.

\subsubsection{Gait--terrain mismatch exposure}
Fraction of time spent (i) trotting while the near band is \textsc{Rough} (trot-on-rough) and (ii) climbing while the near band is \textsc{Smooth} (climb-on-smooth).

\subsubsection{Dwell overhead}
Cumulative time spent in the dwell mode $S$ per run, which captures the stop-and-dwell cost of $T\!\rightarrow\!C$ transitions.

\subsubsection{Traversal time and switching trade-off}
Traversal time from the first to last timestamp of each trial, analyzed jointly with switch count.

\subsubsection{Stability proxy (roll)}
RMS roll angle per trial as a stability proxy.

All metrics are aggregated over trials and summarized in Table~\ref{tab:results_combined}.

%% file: Tables/implementation.tex
\begin{table}[t]
\centering
\caption{Implementation parameters}
\label{tab:impl_params}
\small
\setlength{\tabcolsep}{3pt}
\begin{tabular}{@{}p{0.30\columnwidth} p{0.66\columnwidth}@{}}
\hline
\textbf{Parameter} & \textbf{Value / Description} \\
\hline
Height map grid & $H\times W=128\times128$, $r=0.06$\,m/cell, $Z_{inv} = 10^9$, $N_{min} = 15$\\
Forward wedge & $\theta_{max} = 55^\circ$, $R_{max} = 3.0m$\\
Bands & $K =3$, [0,1), [1,2), [2,3)m\\
Timing &  $\Delta t = 1$, $T_{\mathrm{dwell}} = 1$, $D=1$ \\
Ground patch & $\theta_g = 25^\circ$, $R_g^{min} = 0.10$m, $R_g^{max} = 0.40$m\\ 
Percentiles & $p_h = 95$, $p_s = 90$, $p_d = 5$, $p_g = 10$, $p_T = 90$, $p_C = 40$  \\
Diff Threshold & $\tau = 0.25 $ \\
Reward weights & $\rho_{\text{move}} = 1$, $\rho_{\text{dwell}} = 0.5$, $C_{\text{sw}} = 0.5$, $C_{\text{TR}} = 5$, $C_{\text{CS}} = 1$ \\
\hline
\end{tabular}
\end{table}

%% file: Figures/all_courses.tex
\begin{figure}[t]
    \centering
    \begin{subfigure}[b]{0.32\columnwidth}
        \includegraphics[width=\linewidth]{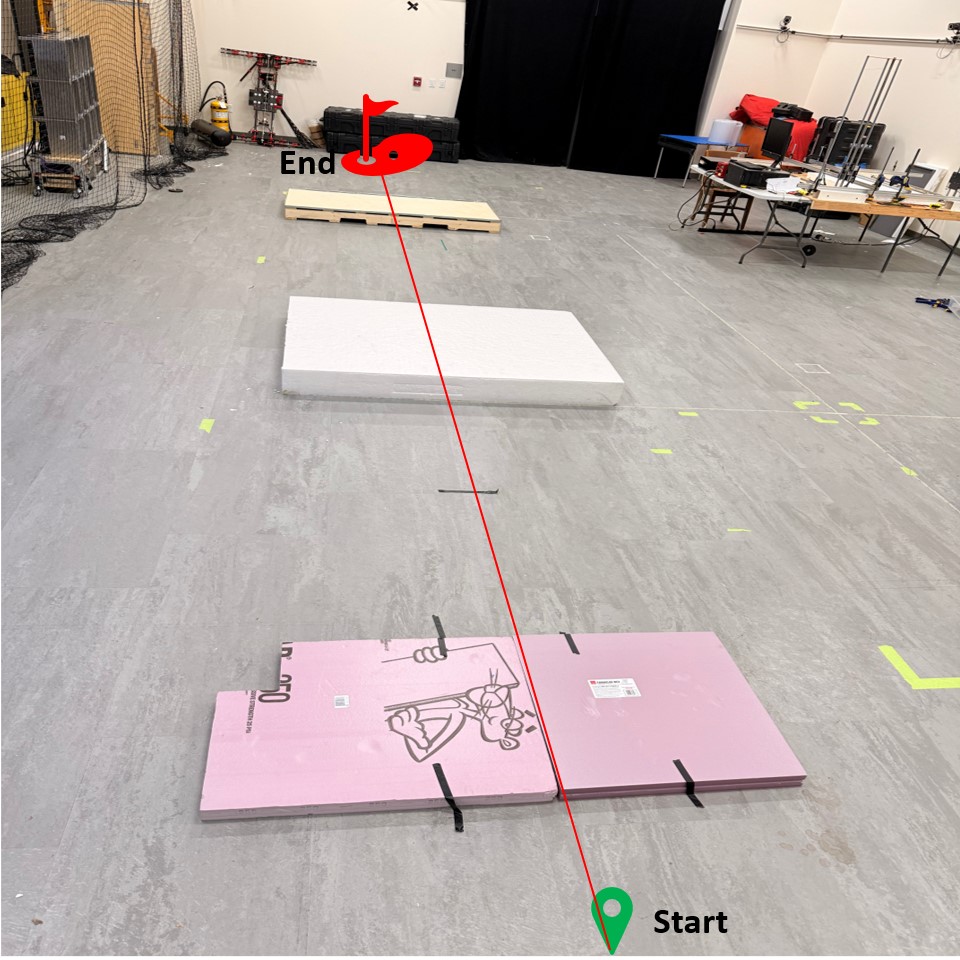}
        \caption{Course 1}
        \label{fig:course1}
    \end{subfigure}
    \hfill
    \begin{subfigure}[b]{0.32\columnwidth}
        \includegraphics[width=\linewidth]{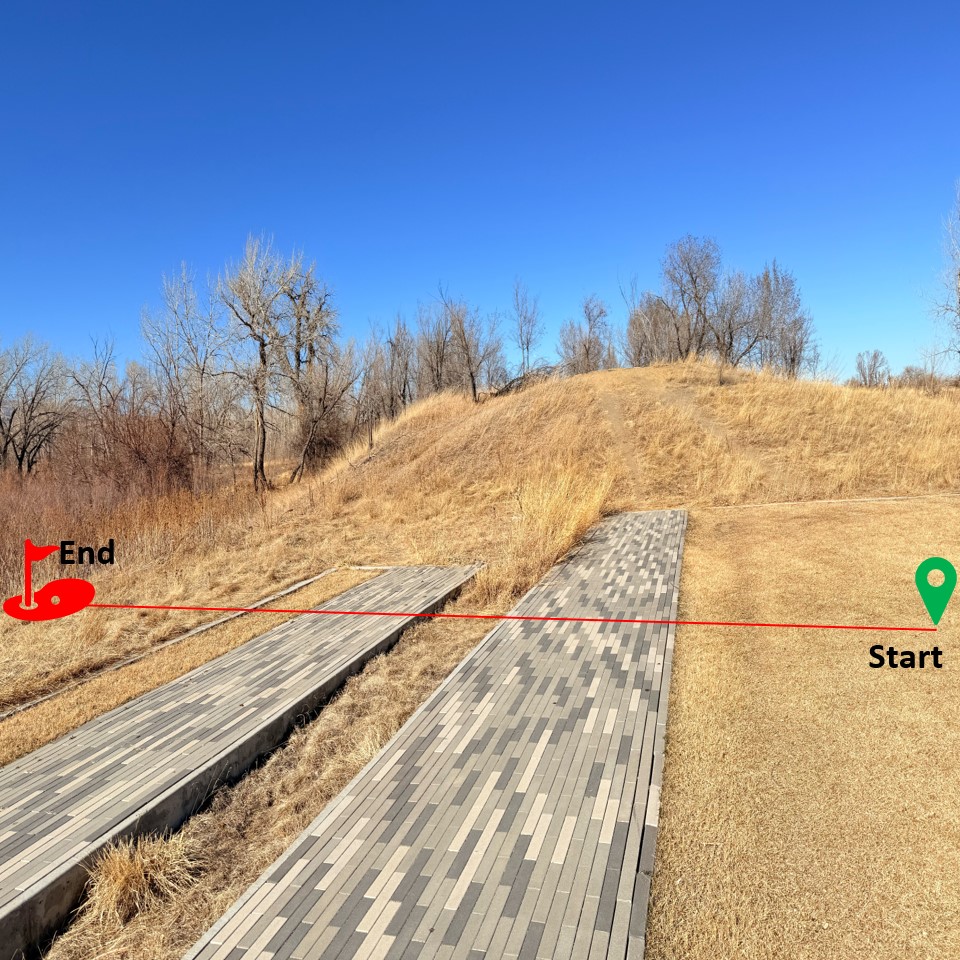}
        \caption{Course 2}
        \label{fig:course2}
    \end{subfigure}
    \hfill
    \begin{subfigure}[b]{0.32\columnwidth}
        \includegraphics[width=\linewidth]{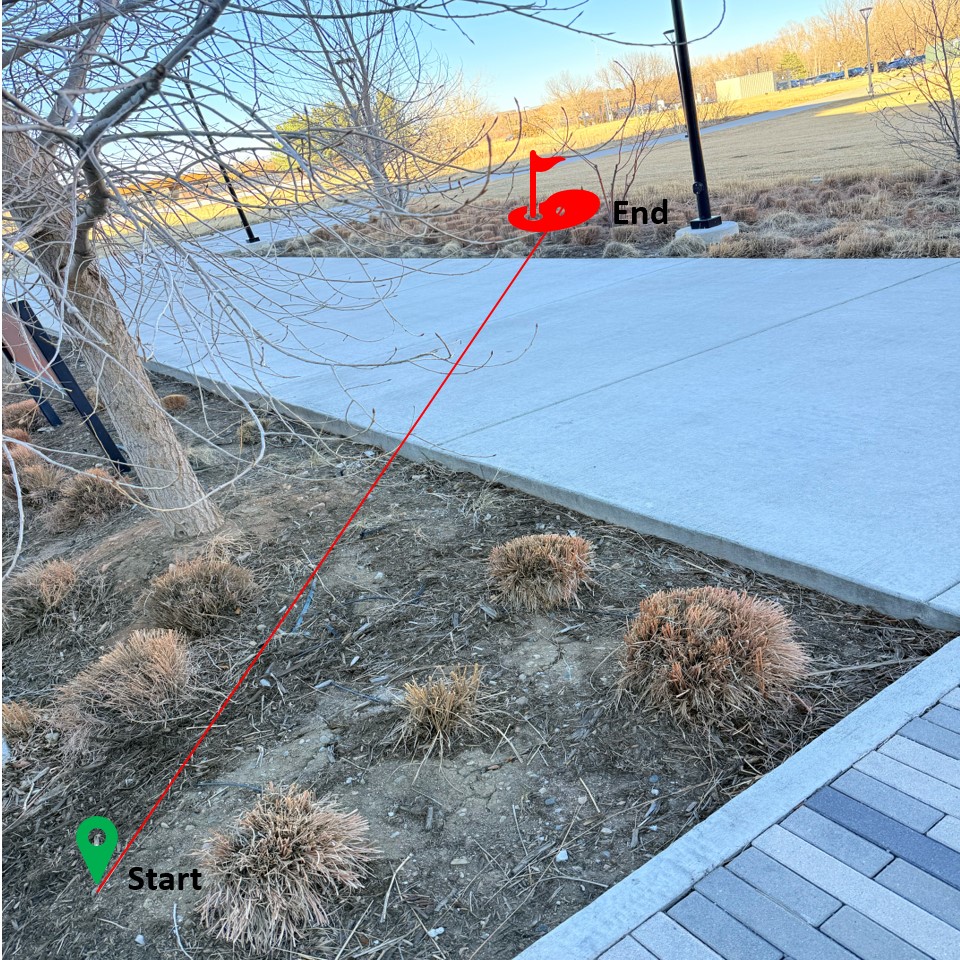}
        \caption{Course 3}
        \label{fig:course3}
    \end{subfigure}
    \caption{Experimental indoor (a) and outdoor (b,c) courses.} 
    \label{fig:courses}
\end{figure}

%% file: Figures/feature_char.tex
\begin{figure}[t!]
    \centering
    \includegraphics[width=\columnwidth]{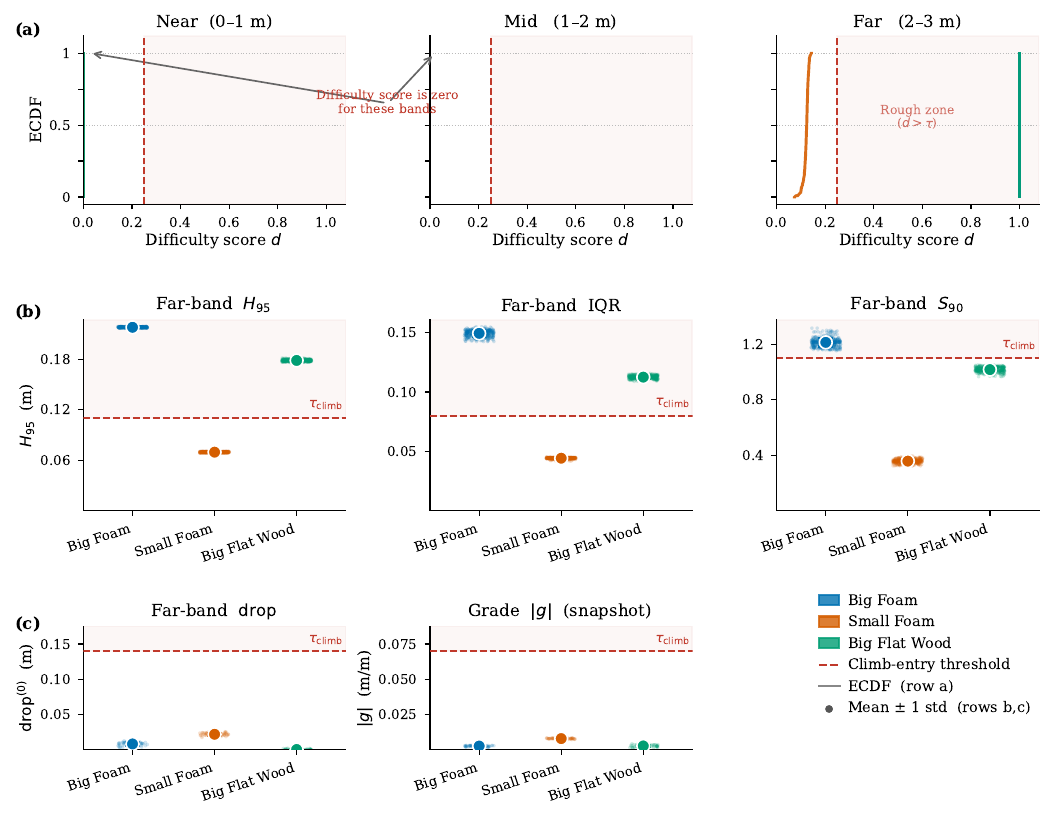}
    \label{fig_lsp_planning}
    \caption{Course 1 feature and look-ahead characterization.}
    \label{fig_feat_dist}
\end{figure}

%% file: sections/results.tex
\input{Tables/summary_course_metric}
\section{Results} \label{section_results}

For each course, we evaluate the three supervisors in Sec.~\ref{subsec:supervisors}. Table~\ref{tab:results_combined} reports per-course metrics (mean $\pm$ 95\% confidence interval). Fig.~\ref{fig:course1_key_metrics} summarizes key metrics for Course~1 using boxplots, and Fig.~\ref{fig:tt_vs_sc_combined} aggregates all runs to visualize the traversal-time versus switch-count trade-off.

\paragraph{Course~1: Indoor Lab Obstacles.}
Fig.~\ref{fig:course1_key_metrics} and Fig.~\ref{fig:course1_aligned_traces} show that LA-MDP initiates $T\!\rightarrow\!C$ earlier relative to near-band rough entry, reflected by a lower Late\,T2C rate and a more negative Lead distance (Fig.~\ref{fig:course1_key_metrics}c). Earlier switching reduces trot-on-rough exposure (Fig.~\ref{fig:course1_key_metrics}b) and improves stability, reflected by lower roll RMS (Fig.~\ref{fig:course1_key_metrics}f). LA-MDP also reduces switch count and dwell overhead (Fig.~\ref{fig:course1_key_metrics}a,d). Traversal time does not necessarily decrease (Fig.~\ref{fig:course1_key_metrics}e), which is consistent with conservative climb occupancy in short rough segments.

\input{Figures/indoor_course}

\paragraph{Course~2: Outdoor Paver Crossing.}
Results are summarized in Table~\ref{tab:results_combined}. Compared to RFT and DCH, LA-MDP reduces trot-on-rough exposure and shifts $T\!\rightarrow\!C$ earlier (lower Late\,T2C and more negative Lead). Traversal time can increase when LA-MDP remains in $C$ to avoid trotting on near-band rough terrain.

\paragraph{Course~3: Outdoor Sidewalk Transition.}
Table~\ref{tab:results_combined} shows that switch count and dwell overhead are similar across methods on this course, indicating that differences are driven primarily by switching timing and gait-terrain mismatch rather than by how often the robot switches. LA-MDP reduces trot-on-rough exposure and improves roll stability through earlier $T\!\rightarrow\!C$ initiation, with a potential traversal-time trade-off.

\paragraph{Cross-course summary.}
Across all three courses, LA-MDP consistently improves $T\!\rightarrow\!C$ timing (lower Late\,T2C and earlier Lead) and reduces gait-terrain mismatch, which correlates with improved roll stability. The main cost is increased traversal time in settings where a conservative preference for $C$ dominates over speed.

%% file: Tables/summary_course_metric.tex
\begin{table*}[t]
\caption{Per-course evaluation metrics (mean $\pm$ 95\% confidence interval, $N=10$ completed runs per method). 
SC: total gait switches; Late\,T2C: fraction of $T\!\rightarrow\!C$ initiated after the near band is already rough; 
ToR: trot-on-rough exposure (time fraction); Lead: median $T\!\rightarrow\!C$ anticipation distance (negative means switch before near-band rough entry); 
Dwell: accumulated stop-and-dwell time; TT: traversal time; Roll\,RMS: time-weighted roll RMS. Bold indicates best per metric per course.}
\centering
\label{tab:results_combined}
\resizebox{\textwidth}{!}{%
\begin{tabular}{|l|l|c|c|c|c|c|c|c|}
\hline
\textbf{Course} & \textbf{Method} & \textbf{SC} & \textbf{Late T2C (\%)} & \textbf{ToR (\%)} & \textbf{Lead (m)} & \textbf{Dwell (s)} & \textbf{TT (s)} & \textbf{Roll RMS ($^\circ$)} \\
\hline
\multirow{3}{*}{\shortstack[l]{Indoor Lab}}
 & RFT    & 4.0$\pm$0.0 & 100.0$\pm$0.0 & 8.2$\pm$1.0 & 0.31$\pm$0.11 & 1.92$\pm$0.02 & 39.1$\pm$1.5 & 2.89$\pm$0.54 \\
 & DCH    & 4.2$\pm$0.4 & 100.0$\pm$0.0 & 6.8$\pm$0.5 & 0.15$\pm$0.03 & 2.01$\pm$0.19 & \textbf{36.4$\pm$1.2} & 2.24$\pm$0.71 \\
 & LA-MDP & \textbf{1.0$\pm$0.0} & \textbf{0.0$\pm$0.0}   & \textbf{0.6$\pm$0.4} & \textbf{$-$0.29$\pm$0.09} & \textbf{1.04$\pm$0.02} & 45.0$\pm$0.4 & \textbf{1.38$\pm$0.11} \\
 \hline
\multirow{3}{*}{\shortstack[l]{Outdoor Paver}}
 & RFT    & 4.0$\pm$0.7 & 100.0$\pm$0.0 & 7.9$\pm$0.8 & 0.28$\pm$0.06 & 2.40$\pm$0.30 & \textbf{48.3$\pm$3.3} & 2.23$\pm$0.20 \\
 & DCH    & 5.4$\pm$0.5 & 96.7$\pm$6.5  & 8.2$\pm$1.5 & 0.07$\pm$0.06 & 3.09$\pm$0.25 & 48.4$\pm$5.1 & 2.03$\pm$0.32 \\
 & LA-MDP & \textbf{2.2$\pm$0.6} & \textbf{30.0$\pm$16.0} & \textbf{1.2$\pm$0.6} & \textbf{$-$0.19$\pm$0.06} & \textbf{1.67$\pm$0.34} & 60.9$\pm$3.1 & \textbf{1.55$\pm$0.25} \\
\hline
\multirow{3}{*}{\shortstack[l]{Outdoor Sidewalk}}
 & RFT    & \textbf{2.8$\pm$0.3} & 100.0$\pm$0.0 & 8.4$\pm$0.9 & 0.25$\pm$0.09 & \textbf{1.75$\pm$0.22} & \textbf{31.3$\pm$1.5} & 3.07$\pm$0.48 \\
 & DCH    & 3.6$\pm$0.8 & 100.0$\pm$0.0 & 7.0$\pm$1.6 & 0.11$\pm$0.09 & 2.22$\pm$0.40 & 33.9$\pm$1.1 & 3.27$\pm$0.37 \\
 & LA-MDP & 3.0$\pm$0.0 & \textbf{55.0$\pm$9.8}  & \textbf{2.5$\pm$0.3} & \textbf{$-$0.32$\pm$0.10} & 2.10$\pm$0.02 & 38.0$\pm$0.9 & \textbf{2.74$\pm$0.32} \\
\hline
\end{tabular}%
}

\label{tab:results_combined}
\end{table*}

%% file: Figures/indoor_course.tex
\begin{figure}[t]
  \centering
  \includegraphics[width=0.9\columnwidth]{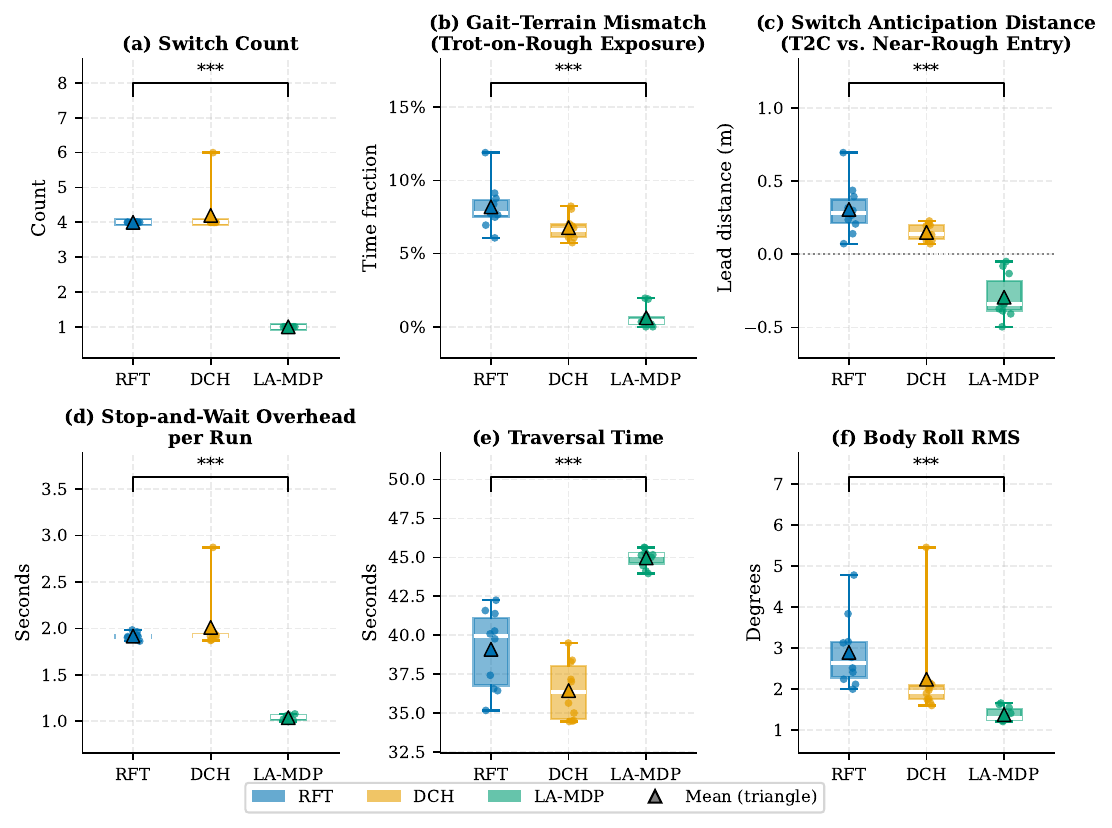}
  \caption{Course~1 (Indoor Lab): key metrics across methods. Significance marker *** indicates a two-sided Mann–Whitney U test with $p < 0.001$.}
  \label{fig:course1_key_metrics}
\end{figure}
\begin{figure}[t]
  \centering
  \includegraphics[width=0.8\columnwidth]{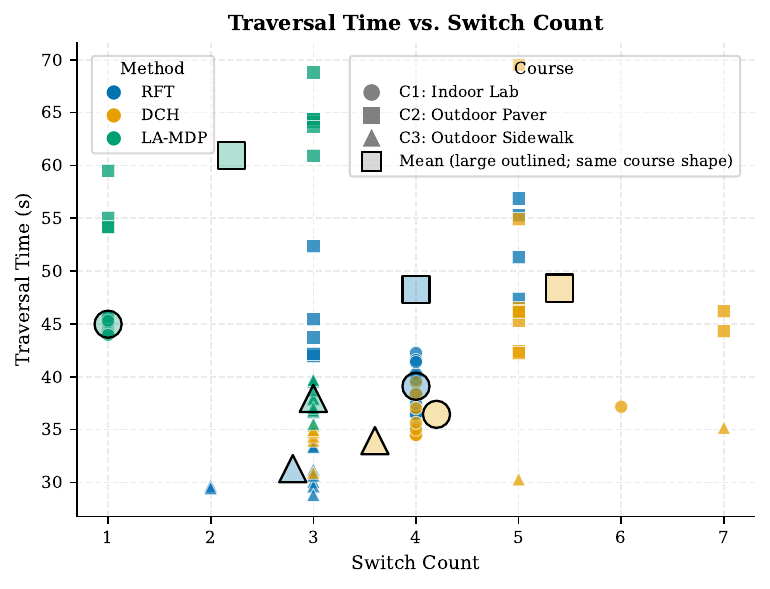}
  \caption{Traversal time vs.\ switch count across all courses.}
  \label{fig:tt_vs_sc_combined}
\end{figure}
\begin{figure}[t]
  \centering
  \includegraphics[width=0.85\columnwidth]{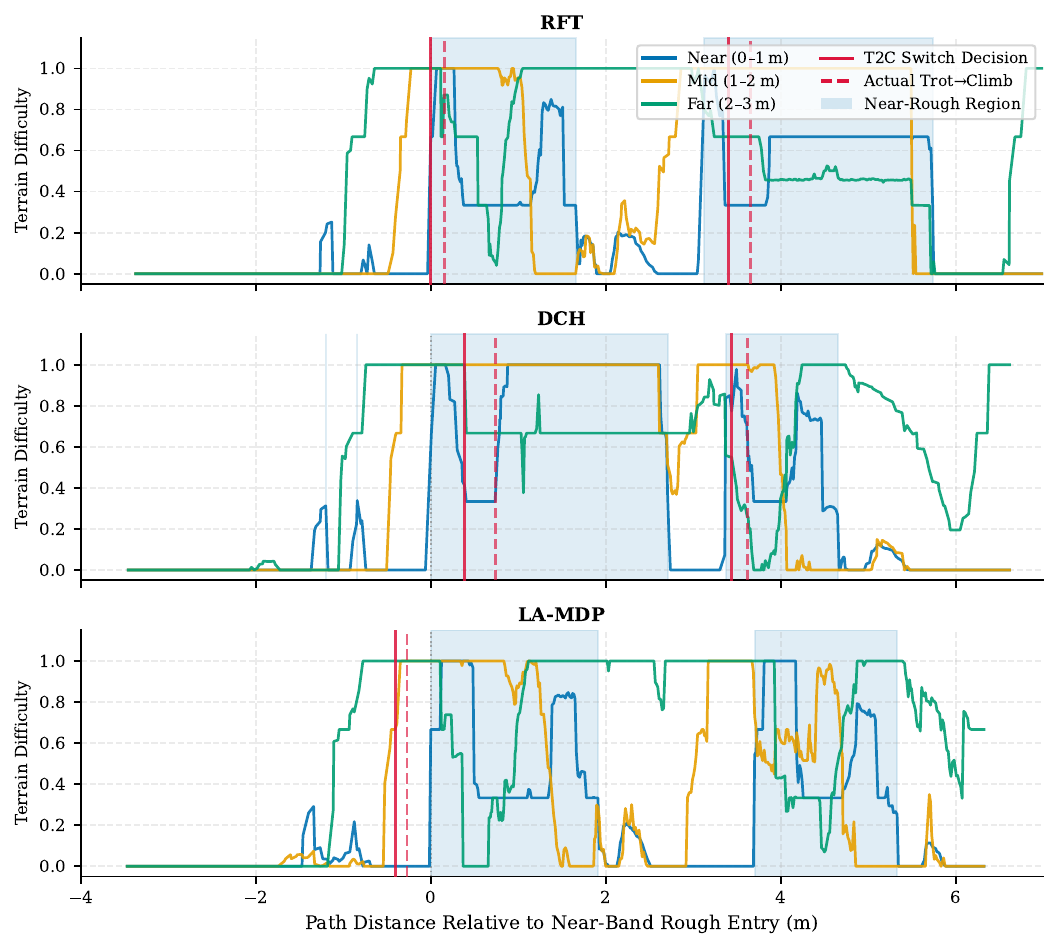}
  \caption{Course~1: Representative distance-aligned difficulty traces with annotated decisions and realized transitions.}
  \label{fig:course1_aligned_traces}
\end{figure}

%% file: sections/conclusion.tex
\section{Conclusions and Future Work} \label{section_conclusion}
Reactive gait switching based on local thresholds cannot reliably anticipate upcoming terrain and therefore tends to incur platform transition costs only after roughness is already imminent. This paper formulates height map-only gait switching as a finite-state, infinite-horizon MDP over a multi-band look-ahead representation derived from a robot-centered 2.5D height map. The key idea is to model asymmetric transition behavior directly in the MDP state via an explicit dwell mode, so the policy accounts for stop-and-dwell latency as part of the decision process rather than as an external heuristic.

Experiments on a Unitree Go2 across three indoor and outdoor courses show that the look-ahead MDP policy reduces late $T\!\rightarrow\!C$ initiations, decreases gait--terrain mismatch exposure, and improves roll stability relative to reactive baselines. These robustness gains come with a traversal-time trade-off in settings where conservative climb occupancy dominates speed. Overall, the results support the principle that platform-specific locomotion constraints should be represented explicitly in the decision model used for gait switching.

Future work will incorporate explicit efficiency objectives (time and energy) into the reward to better manage robustness--speed trade-offs. We also plan to move beyond binary \textsc{Smooth}/\textsc{Rough} classes using richer height map cues and to expand the action set to additional locomotion primitives and turning behaviors for terrain-aware navigation.